Original article

# Generative augmentation of imbalanced flight records for flight diversion prediction: A multi-objective optimisation framework

Karim Aly *, Alexei Sharpanskykh, Jacco Hoekstra

*Control and Operations, Faculty of Aerospace Engineering, Delft University of Technology (TU Delft), Kluyverweg 1, Delft, 2629 HS, South Holland, Netherlands*

A B S T R A C T

Flight diversions are rare but high-impact events in aviation, making their reliable prediction vital for both safety and operational efficiency. However, their scarcity in historical records impedes the training of machine learning models used to predict them. This study addresses this challenge by proposing a generative augmentation framework for imbalanced aviation tabular records. The principal contribution lies in the design of a composite optimisation objective specifically tailored to flight data, which integrates four complementary quality dimensions into a single score used to guide automated hyperparameter search via the Tree-structured Parzen Estimator (TPE) algorithm: realism, statistical similarity, fidelity, and predictive utility. These dimensions were selected and defined to reflect the operational and statistical requirements specific to aviation records, and were complemented by two descriptive evaluation dimensions, diversity and operational validity, forming a six-stage assessment framework. The composite objective was then used to tune three deep generative models, namely Tabular Variational Autoencoder (TVAE), Conditional Tabular Generative Adversarial Network (CTGAN), and CopulaGAN, with Gaussian Copula (GC) serving as a statistical baseline. Results show that optimised models substantially outperform their default counterparts across all six assessment dimensions, and that augmentation with the resulting synthetic data improves diversion prediction compared to training on real data alone. These findings demonstrate that domain-adapted multi-objective optimisation is an effective strategy for generative augmentation of rare events in aviation, with applicability to other imbalanced tabular prediction tasks.

## 1. Introduction

Flight diversions represent critical disruptions in air transportation operations. A diversion typically occurs when an aircraft cannot complete its journey to the planned destination airport and must land at an alternate airport. Such events are costly for airlines, which face increased fuel expenses, crew rescheduling, and potential compensation to passengers [1]. Diversions also impose significant burdens on passengers, who experience delays, missed connections, and inconvenience, as well as on airports and air traffic controllers, who must quickly adapt operational plans to accommodate unplanned arrivals [2]. Beyond economic and operational impacts, diversions can have safety implications, particularly when they are driven by adverse weather, technical malfunctions, or medical emergencies on board.

Given these wide-ranging consequences, reliable prediction of diversions is essential [3]. Accurate forecasts can support proactive decision-making, allowing airlines and air traffic management to anticipate and mitigate disruptions. For instance, predictive tools could help identify flights at risk of diversion before departure or during en-route monitoring, enabling operators to adjust flight plans, allocate resources more effectively, and minimise downstream operational effects. Improved predictive capability is therefore closely tied to enhancing both the efficiency and resilience of the air transport system.

While artificial intelligence (AI) is increasingly adopted across the aviation domain to optimise operations and support decision-making, predicting diversions still poses unique challenges due to their rarity in historical records. Flight diversions account for only a small fraction of total flights, making them a classic case of a rare event in machine learning [4]. From a data perspective, this leads to severe class imbalance, where the majority class (non-diverted flights) vastly outnumbers the minority class (diverted flights). Class imbalance can bias learning algorithms toward the majority class, resulting in models that achieve high overall accuracy but fail to correctly identify diversion events. Consequently, the scarcity of diversion cases in the data hinders the development of robust and reliable predictive models.

A further complication is that publicly available flight data often lacks features strongly correlated with diversions. While operational records include variables such as departure delays, weather conditions, and airport characteristics, these attributes may not capture the complex causal factors that ultimately trigger diversions. The weak correlation between observed features and diversion outcomes further limits the ability of conventional machine learning (ML) models to discriminate effectively between diverted and non-diverted flights.

* Corresponding author.
*E-mail addresses:* k.y.s.b.aly@tudelft.nl (K. Aly), o.a.sharpanskykh@tudelft.nl (A. Sharpanskykh), j.m.hoekstra@tudelft.nl (J. Hoekstra).

Synthetic data augmentation offers a promising pathway to address these challenges. By generating artificial flight records that mimic the statistical properties of real diversions, generative models can enrich the minority class and alleviate class imbalance. Augmenting training datasets with realistic synthetic diversion cases provides machine learning models with a broader and more representative sample of rare events, improving their ability to generalise and detect true diversions in unseen data. In this way, synthetic data techniques open new opportunities for advancing the predictive modelling of rare but high-impact events in aviation, with the potential to improve both safety and operational efficiency.

Although generative approaches such as Gaussian Copula (GC), Tabular Variational Autoencoder (TVAE), Conditional Tabular Generative Adversarial Network (CTGAN), and CopulaGAN [5,6] have demonstrated the ability to produce realistic tabular data, their application to highly imbalanced flight records is largely unexplored. This study does not propose a new generative architecture; rather, it investigates the applicability of these established models—including the recent GAN-based tabular methods CTGAN and CopulaGAN—to the specific challenges posed by rare aviation events. These models are sensitive to hyperparameter choices, especially when trained on very small datasets, which is often the case for rare events like diversions. Moreover, aviation data present additional challenges due to temporal dependencies, mixed feature types, and operational constraints. Addressing these issues requires careful adaptation of generative architectures and systematic hyperparameter optimisation to ensure the production of realistic, high-quality synthetic flight records.

This study investigates whether generative models can enhance flight diversion prediction under conditions of extreme class imbalance (127 diversions out of 61,000 flights). The novel contributions of this research are fourfold: (i) a multi-objective optimisation framework for tuning generative model hyperparameters on rare flight events, (ii) demonstrating that the assessment of synthetic data quality requires a comprehensive, multi-faceted evaluation rather than reliance on a single metric, (iii) a comparative evaluation of predictive accuracy between models trained exclusively on real data and those trained on a combination of real and synthetic data, and (iv) a demonstration of how different augmentation sizes affect the quality of predicting rare events such as flight diversions. By this approach, we provide new evidence on the role of synthetic augmentation in improving the prediction of rare aviation events under conditions of extreme class imbalance.

To achieve these goals, we train and optimise multiple generative models on the diversion subset, generate synthetic diversion records, and assess data quality using a six-stage evaluation framework encompassing realism, diversity, operational validity, statistical similarity, fidelity, and predictive performance. Rather than identifying a single best generative model or proposing a novel architecture, the aim is to demonstrate how our multi-objective hyperparameter optimisation improves synthetic data quality compared to non-optimised baselines, and to quantify the contribution of synthetic augmentation to predictive modelling of flight diversions.

The remainder of this paper is organised as follows: Section 2 reviews related work, Section 3 presents the augmentation and optimisation framework, Section 4 reports experimental results, Section 5 provides insights and limitations, and Section 6 concludes with key findings and future directions.

## 2. Related work

While synthetic data has gained widespread adoption across diverse domains, including finance, healthcare, cybersecurity, computer vision, and manufacturing, its application in air transportation remains relatively underexplored. Current research in aviation analytics has predominantly focused on downstream machine learning tasks, utilising historical data to predict operational disruptions such as delays, cancellations, diversions, and turnaround times. However, these datasets typically exhibit severe class imbalance, with critical events like flight diversions occurring substantially less frequently than routine operations, thereby creating significant challenges for accurate diversion prediction. Studies on related aviation tasks, such as flight delay prediction, have similarly confronted this imbalance challenge. For instance, Khan et al. [7] combined sampling techniques, including SMOTETomek, with a hierarchical machine learning model to predict departure delay status and duration from highly skewed airline data, while Khan et al. [8] evaluated fourteen sampling approaches in conjunction with a parallel-series model for IATA-coded delay subcategory prediction.

Applications of synthetic data within aviation have primarily focused on trajectory generation tasks. Representative studies include synthetic trajectory generation for routing optimisation in Aeronautical Ad-hoc Networks (AANETs) [9], trajectory reconstruction using Large Language Models (LLMs) [10], and the generation of less frequent landing patterns in the Terminal Manoeuvring Area (TMA)—such as go-arounds and holding trajectories—through Time Generative Adversarial Networks (TimeGAN) [11]. In addition, Murad and Ruocco [12] introduced an end-to-end trajectory synthesis approach based on the Time-Based Vector Quantised Variational Autoencoder (TimeVQVAE). Beyond trajectory-focused applications, synthetic data has also been employed to address class imbalance: for instance, Ahmadi et al. [13] combined generative methods with GPT-5 to augment text-based accident datasets, thereby reducing imbalance and enhancing reinforcement learning classification performance. More directly relevant to this study, our earlier work [14] demonstrated the potential of generative models for producing comprehensive synthetic flight records—encompassing flight identifiers, airline and aircraft information, origin-destination pairs, scheduled and actual timestamps, air time, and operational logs of delays, cancellations, and diversions—providing the foundation that this paper extends by specifically targeting the diversion subset to augment the minority class.

Flight diversion prediction has attracted comparatively limited research attention. Di Ciccio et al. [2] proposed one of the earliest dedicated diversion prediction models, designed for freight logistics providers requiring early warning of in-flight diversions. They derived behavioural features from real-time flight tracking data, specifically progress toward origin and destination, velocity deviation, and altitude deviation over time intervals, and trained a one-class SVM on non-diverting flights to detect anomalous behaviour. A diversion was predicted when the number of consecutive anomalous intervals exceeded a threshold. Evaluated on 1,130 flight tracks, of which 68 were diverted, the model achieved a precision of 0.91, a recall of 0.74, and an F-score of 0.82, providing an average response time gain of 62 min relative to the actual landing time. This approach, however, requires real-time trajectory data and cannot produce predictions before departure. Li et al. [15] addressed en-route diversion prediction from a weather avoidance perspective, training an SVM on historical flight and weather data with PCA used to reduce ten weather indicators; the model outperformed k-nearest neighbour, logistic regression, random forest, and deep neural network baselines in accuracy, precision, and F1. Dalmau and Gawinowski [16] adopted a pre-departure approach, combining flight plan attributes, METAR/SPECI weather observations, and ATFM regulation data as input to a gradient-boosted decision trees model trained on 13.6 million flights in the IFPS zone, of which only 0.2% were diverted. Confident learning was applied to prune training observations likely diverted for non-weather reasons, such as medical emergencies. The model achieved a precision of 0.82 and a recall of 0.32, with destination airport, aircraft type, and weather phenomena identified as the most influential predictors. Predictive performance degraded substantially with increasing look-ahead time, as the model depended on weather observa-tions near the estimated time of arrival; at four hours before arrival, average precision dropped to 0.08. Dalmau and Gawinowski [3] extended this work by applying supervised clustering, combining Shapley value attribution, UMAP dimensionality reduction, and HDBSCAN, to characterise the conditions under which the model succeeded or failed in predicting diversions during a live trial. Six diversion clusters were identified: restrictive minima, extreme obscuration, storms, thunderstorms, basic aircraft equipment, and strong winds, with recall ranging from 0.70 for extreme obscuration to 0.00 for storm-related and equipment-related diversions. Across all of these studies, however, class imbalance was treated as an inherent characteristic of the data rather than a problem to be addressed through augmentation.

The only study to have addressed this imbalance challenge in diversion prediction reveals further limitations. Buddhadev [17] applied the Synthetic Minority Oversampling Technique (SMOTE) to balance airline datasets for diversion prediction. However, the resulting models exhibited signs of overfitting, exposing limitations of interpolation-based oversampling approaches that do not explicitly account for data realism or operational validity. Beyond interpolation-based methods, deep generative models such as TVAE and CTGAN have also been shown to produce substantially lower-quality minority-class samples when trained on imbalanced tabular datasets [18]. In addition, several studies reported that these models often underperform compared to simpler augmentation approaches when used with default configurations, highlighting the strong sensitivity of synthetic tabular data quality to hyperparameter selection [19–21]. Despite this, most prior work either evaluates synthesisers using default settings or optimises only a limited subset of quality criteria, typically focusing on realism alone [21] or on restricted combinations of fidelity and utility metrics in privacy-oriented settings [20]. Consequently, an important research gap remains: existing studies have not investigated how to jointly optimise multiple complementary quality dimensions specifically for minority-class augmentation in aviation tabular records. To the best of our knowledge, no prior work has applied multi-objective hyperparameter optimisation to generative models for synthetic aviation diversion records, nor evaluated the generated data across the full set of quality dimensions required for operationally reliable augmentation.

To address this gap, the proposed framework combines deep learning-based generative models with a multi-objective optimisation strategy tailored to aviation tabular data, benchmarked against a statistical baseline. Specifically, the framework formulates four complementary quality dimensions—realism, statistical similarity, fidelity, and predictive utility—as a composite optimisation objective, embedded within a broader six-stage evaluation framework that additionally assesses diversity and operational validity. This design directly addresses the two limitations identified above: the absence of domain-adapted quality criteria for aviation tabular records, and the lack of a multi-objective optimisation strategy to enforce them jointly during synthesis.

## 3. Methodology

This section outlines the methodological framework, which is organised into three sequential modules, as illustrated in Fig. 1. Module 1 (Data Preparation) collects approximately 61,000 U.S. domestic flight records from the TranStats database, applies preprocessing and feature engineering, and produces a set of preprocessed real flight records—comprising both normal and diverted flights—that serve as input to the subsequent modules. Module 2 (Synthetic Data Generation) takes as input only the 127 diverted records, trains four generative models on this minority class, and optimises their hyperparameters via a composite objective function targeting realism, statistical similarity, fidelity, and predictive utility, producing a set of synthetic flight diversions. Module 3 (Quality Evaluation) operates on four inputs: the synthetic diversions from Module 2; the real diversion records from Module 1; the full real flight records from Module 1; and the augmented records formed by combining the full real records with the synthetic diversions. These inputs are used to evaluate synthetic data quality across six complementary dimensions and to assess the utility of augmentation for diversion prediction through a Train-Augmented-Test-Real experimental protocol.

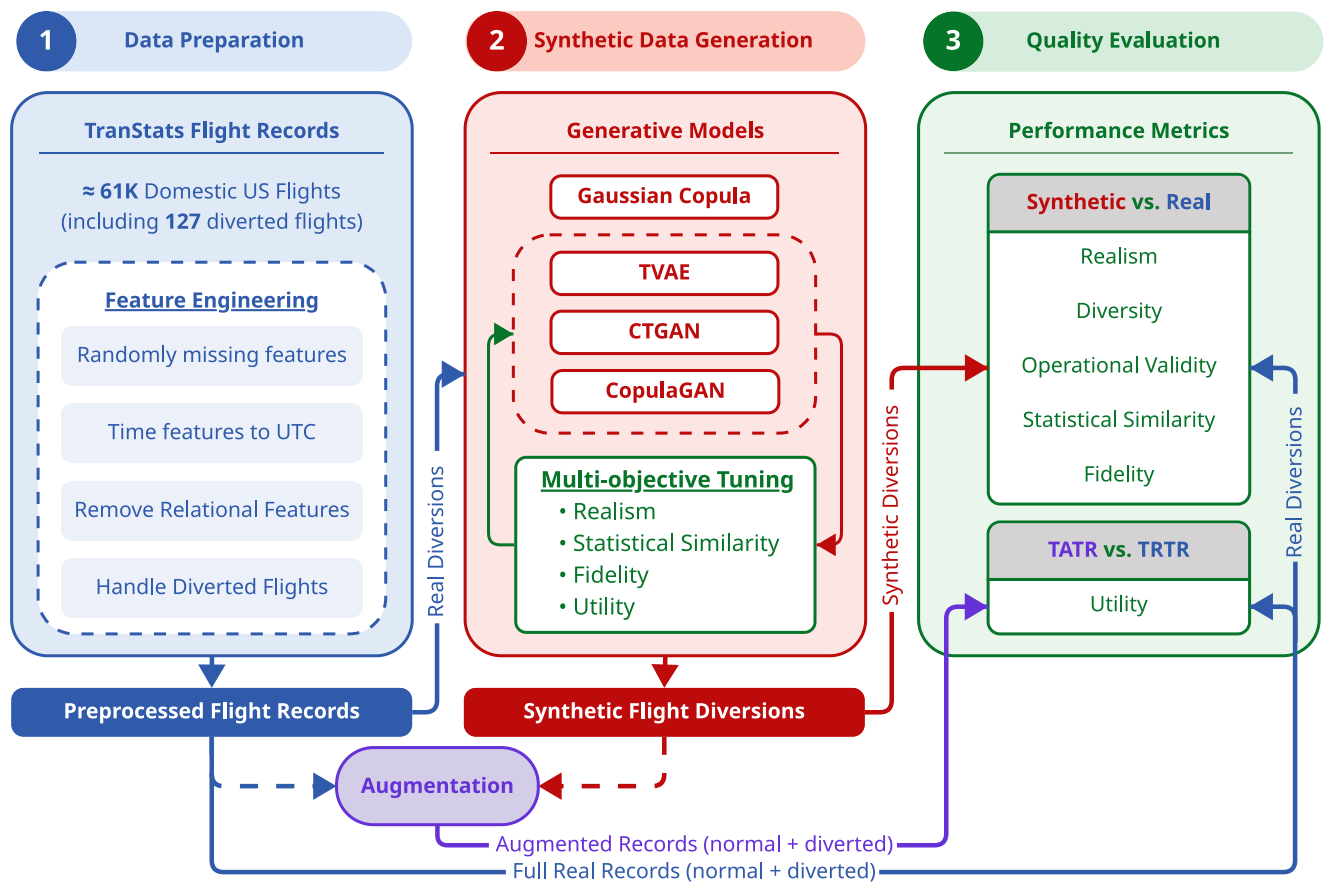

**Fig. 1.** Overview of the analysis framework. Module 1 (Data preparation): Collection, preprocessing, and feature engineering of approximately 61,000 U.S. domestic flight records. Module 2 (Synthetic data generation): Four generative models tuned via multi-objective hyperparameter optimisation on the 127 diverted flights. Module 3 (Quality evaluation): Assessment of the generated records across six complementary dimensions and evaluation of their utility for diversion prediction.

### 3.1. Data and preprocessing

This study relied on the publicly available “TranStats Database for Airline On-Time Performance” provided by the Bureau of Transportation Statistics (BTS) [22,23]. The database covers U.S. domestic flights and contains comprehensive records on delays, cancellations, diversions, and their causes. Its richness makes it an invaluable source for modelling a wide range of use cases in air transport research.

The dataset, preprocessing, and feature engineering were largely identical to our earlier work [14]. In this analysis, 14 features were selected to train the generative models, while the remaining attributes—hereafter referred to as relational features—were inferred post-generation to maximise realism and preserve feature dependencies. Table 1 lists the features used for training the generative models and the subset employed for the diversion prediction task described in Section 3.4.

In the historical records, diverted flights could be easily distinguished from non-diverted ones. For instance, diverted flights lacked values for features such as “Actual Elapsed Time (min)” and “Air Time (min)”, while including diversion-specific attributes such as “Diversion Actual Elapsed Time (min)” and “Diversion Distance (miles)”. Since our objective was to train a classifier to predict diversions, it was essential to ensure diverted and non-diverted flights were represented with the same feature set; otherwise, the classification task would have been trivial. To achieve this, the missing “Actual Elapsed Time (min)” was imputed using “Diversion Actual Elapsed Time (min)”. The “Air Time (min)” was then reconstructed by subtracting “Taxi In Time (min)” and “Taxi Out Time (min)” from this value. Finally, all remaining diversion-specific attributes were removed prior to training the prediction model.

### 3.2. Generative models

In this study, we employed four models to generate synthetic records of diverted flights: Gaussian Copula (GC), Tabular Variational Autoencoder (TVAE), Conditional Tabular Generative Adversarial Network (CTGAN), and CopulaGAN. Since diverted flights were extremely rare compared to non-diverted ones, the objective was to augment this minority class with synthetic samples in order to construct a more balanced dataset suitable for training diversion prediction models. This can be seen as "highlighting" rare events by increasing their frequency in the data, thereby improving the training and predictive performance of the models.

**Table 1**
Features used in this analysis, categorised as: included (✓), excluded (✗), or calculated post-generation (🖩).

| Features | for generation | for prediction |
|---|---|---|
| Unique Carrier Code | ✓ | ✓ |
| Tail Number | ✓ | ✓ |
| Origin Airport ID | ✓ | ✗ |
| ICAO Origin Airport | 🖩 | ✓ |
| Origin City | 🖩 | ✗ |
| Origin State Code | 🖩 | ✗ |
| Origin State Name | 🖩 | ✗ |
| Destination Airport ID | ✓ | ✗ |
| ICAO Destination Airport | 🖩 | ✓ |
| Destination City | 🖩 | ✗ |
| Destination State Code | 🖩 | ✗ |
| Destination State Name | 🖩 | ✗ |
| Quarter | 🖩 | ✓ |
| Day of Week | 🖩 | ✓ |
| Scheduled Departure Time UTC | ✓ | ✓ |
| Actual Departure Time UTC | ✓ | ✓ |
| Departure ΔT (min) | ✓ | ✓ |
| Departure Delay Label | 🖩 | ✗ |
| Taxi Out Time (min) | ✓ | ✓ |
| Wheels Off Time UTC | 🖩 | ✓ |
| Wheels On Time UTC | 🖩 | ✗ |
| Taxi In Time (min) | ✓ | ✗ |
| Scheduled Arrival Time UTC | 🖩 | ✓ |
| Actual Arrival Time UTC | 🖩 | ✗ |
| Arrival ΔT (min) | ✓ | ✗ |
| Arrival Delay Label | 🖩 | ✗ |
| Scheduled Elapsed Time (min) | ✓ | ✓ |
| Actual Elapsed Time (min) | ✓ | ✗ |
| Air Time (min) | ✓ | ✗ |
| Distance (miles) | 🖩 | ✓ |
| Diversion Label | ✓ | **Target** |

The four models were selected to provide methodological diversity across the main generative paradigms established for tabular data. GC is a statistical model that estimates distributions directly from the data without relying on a deep architecture, and serves as a non-deep baseline. TVAE and CTGAN represent two well-established deep learning paradigms—variational autoencoding and conditional adversarial training, respectively—and were included for their complementary strengths. TVAE offers stable training and reliable convergence, as VAE-based models are less susceptible to mode collapse than GANs, which is a particularly relevant property given the small size of the minority class. CTGAN, by contrast, introduces a conditional sampling mechanism that explicitly targets underrepresented categorical values during training, making it well-suited to mixed-type tabular data with class imbalance, at the cost of greater training sensitivity. CopulaGAN occupies a hybrid position: it applies copula-based preprocessing to the data before adversarial training, combining the dependence-modelling properties of GC with the generative capacity of CTGAN. Including all four models allows the evaluation framework to be tested across fundamentally different inductive biases, and ensures that any improvements attributable to multi-objective optimisation are not specific to a single generative approach. All four models were trained on the 127 real diversion records and evaluated under both default hyperparameter configurations [5] and configurations optimised using our multi-objective tuning framework, with the exception of GC, which served as a fixed statistical baseline.

GC is a statistical approach that models complex dependencies among variables while preserving their marginals [24]. It achieves this by transforming each marginal distribution into a uniform distribution through its cumulative distribution function (CDF) and combining them via a Gaussian copula [25–27]. The dependence structure is defined by the correlation matrix of the underlying multivariate normal distribution. For random variables $X_1, X_2, \dots, X_d$ with marginals $F_1(x_1), F_2(x_2), \dots, F_d(x_d)$, the copula $C_\theta$ is expressed as:

$$C_\theta\big(F_1(x_1), F_2(x_2), \dots, F_d(x_d)\big) = \Phi_\theta\big(\Phi^{-1}(F_1(x_1)), \Phi^{-1}(F_2(x_2)), \dots, \Phi^{-1}(F_d(x_d))\big) \quad (1)$$

where $\Phi_\theta$ denotes the joint CDF of the multivariate normal distribution with correlation matrix $\theta$, and $\Phi^{-1}$ is the inverse CDF of the standard normal distribution. Rather than assuming a fixed parametric form for the marginals, we combined Gaussian copulas with Kernel Density Estimation (KDE) from Synthetic Data Vault [28]. A Gaussian kernel was employed to estimate each marginal distribution, after which the copula model was used to encode interdependencies. This approach allowed the generated samples to reflect both accurate marginal behaviour and joint dependencies of the real data.

The TVAE is a deep generative model that extends conventional autoencoders by incorporating probabilistic modelling tailored to tabular datasets [29]. It consists of an encoder that maps input data $x$ into a distribution over a latent space $z$, denoted as $q(z|x)$, and a decoder that reconstructs the input as the conditional distribution $p(x|z)$. Training is performed by maximising the Evidence Lower Bound (ELBO) on the log-likelihood of the observed data, denoted by:

$$ELBO = \mathbb{E}_{z\sim q(z|x)}[\log p(x|z)] - D_{\mathrm{KL}}(q(z|x)\|p(z)) \quad (2)$$

where the first term encourages accurate data reconstruction, and the second minimises the Kullback–Leibler divergence between the approximate posterior $q(z|x)$ and the prior $p(z)$, ensuring a well-structured latent space [30–32].

The CTGAN [6] is a GAN-based architecture tailored for tabular data, designed to handle mixed data types and imbalanced categorical distributions. Like other GANs [33], it consists of a generator $G$ and a discriminator $D$ trained adversarially. CTGAN introduces two key innovations: (i) a conditional sampling mechanism that allows the generator to focus on underrepresented categorical values during training, and (ii) mode-specific normalisation that transforms continuous variables into Gaussian mixture distributions, enabling effective modelling of non-Gaussian and multimodal features. The adversarial training follows the standard minimax loss:

$$\min_G \max_D V(D, G) = \mathbb{E}_{x\sim p_{\mathrm{data}}(x)}[\log D(x)] + \mathbb{E}_{z\sim p_z(z)}[\log(1 - D(G(z)))] \quad (3)$$

where $p_{\mathrm{data}}(x)$ represents the real data distribution and $p_z(z)$ the prior (typically Gaussian) from which generator inputs are sampled. The conditioning on categorical variables during training forces the generator to capture realistic relationships across both majority and minority classes, making CTGAN particularly effective for imbalanced data.

CopulaGAN [28] combines copula transformations with adversarial learning to capture dependencies in tabular datasets. Each feature is first transformed to a uniform distribution using its empirical CDF, and a Gaussian copula is applied to model dependencies among variables. The generator and discriminator are then trained on this copula-transformed space using the adversarial objective in (3). Finally, an inverse copula transformation maps the synthetic samples back to the original feature space, restoring the observed marginals. Unlike CTGAN, which handles imbalance via conditional sampling, CopulaGAN leverages statistical copula transformations to represent complex dependencies before adversarial training.

### 3.3. Experiments

Seven experiments were conducted to systematically assess the impact of the multi-objective hyperparameter optimisation on the quality of the generated datasets. The structure of these experiments is presented in Table 2.

**Table 2**
Generative experiments.

| Experiment | Model used for generation |
|---|---|
| $GC$ | Gaussian Copula model with Kernel Density Estimation |
| $TVAE$ | Tabular Variational Autoencoder model |
| $TVAE_{opt.}$ | Optimised Tabular Variational Autoencoder model |
| $CTGAN$ | Conditional Tabular GAN model |
| $CTGAN_{opt.}$ | Optimised Conditional Tabular GAN model |
| $CopulaGAN$ | CopulaGAN model |
| $CopulaGAN_{opt.}$ | Optimised CopulaGAN model |

Table 3 reports the size of real diversion records used as input to train generative models, along with the synthetic samples drawn from the learned distributions. After generation, relational features were reconstructed, and rejection sampling was applied to remove routes not present in the historical data. The resulting cleaned datasets were then evaluated using the six-stage framework described in the following section.

**Table 3**
Data sizes.

| Stage | Data size |
|---|---|
| Input (real) | (127, 14) |
| Sampled (syn.) | (1000, 14) |
| Reconstructed | (1000, 31) |
| Cleaned | Varies with models |

### 3.4. Evaluation framework

Robust validation is indispensable, as synthetic data that lacks credibility can undermine the very analyses it aims to support. An essential validation step ensures that synthetic data preserves the real structure—matching the number of features, keeping continuous values within observed ranges, and constraining discrete values to their original categories. Building on these checks, our framework assesses six complementary dimensions:

#### 3.4.1. Realism assessment

Realism evaluates whether synthetic routes reflect historical connections. Novel origin–destination pairs not observed in real data are considered invalid, representing broken relations learned by the generative models. The realism score is defined as the percentage of valid routes, with higher values indicating better performance. Invalid routes are removed before subsequent evaluations.

#### 3.4.2. Diversity assessment

Diversity was evaluated using dimensionality reduction applied to both real and synthetic datasets. Principal Component Analysis (PCA) [34] and t-distributed Stochastic Neighbour Embedding (t-SNE) [35] were used to project the data into two dimensions, enabling visual comparison of distributions and clusters. PCA captures linear relationships, whereas t-SNE preserves non-linear structures. In addition, class balance was inspected by comparing the proportions of on-time and delayed departures, ensuring alignment with historical records. Preserving diversity is vital, as insufficient variability risks producing biased or unrepresentative downstream models.

#### 3.4.3. Operational validity assessment

Operational validity was assessed by comparing the correlation between key operational attributes, specifically “Air Time (min)” and “Distance (miles)”, in real versus synthetic data. Implausible records—e.g., extremely short flight times for long routes—indicate violations of operational feasibility in the synthetic dataset. While this study focused on this single correlation, additional operational relationships are expected to be incorporated in future work.

#### 3.4.4. Statistical assessment

Statistical similarity was evaluated through both visual and quantitative analyses. Visual inspections compared marginal and bivariate distributions, while quantitative measures included the Kolmogorov–Smirnov test for numerical and datetime features [36], and Total Variation Distance for categorical and boolean variables [37]. Pairwise relationships were further assessed using Correlation Similarity [38] for numerical features and Contingency Similarity [39] for categorical features. Combining univariate and pairwise assessments provides a comprehensive evaluation of distributional similarity, avoiding misleading conclusions from relying solely on univariate checks.

#### 3.4.5. Fidelity assessment

Fidelity was tested by training a binary classifier to discriminate between real and synthetic samples. A Random Forest model was selected for its ability to capture complex, non-linear interactions [40]. Stratified K-fold cross-validation with five splits and shuffling was applied to maintain class proportions and enhance robustness [41]. Because synthetic diversions outnumbered real diversions, performance was reported using the F1 score and balanced accuracy [42], the latter mitigating class imbalance by weighting minority and majority classes equally. A lower classification accuracy indicates greater fidelity, as the classifier struggles to separate synthetic from real data.

While ensuring that synthetic records are indistinguishable from real ones is a necessary condition for high fidelity, it is equally important to verify that they are not near-copies of real training samples. To this end, the Distance to Closest Record (DCR) was computed for each synthetic sample [43] as a direct test of memorisation. DCR is defined as the Euclidean distance from a synthetic record to its nearest neighbour in the real dataset, computed on numerical features in a standardised feature space using $z$-score normalisation fitted to the real data. To establish a reference, a real-to-real baseline was computed as the distance from each real record to its nearest *other* real sample (i.e., the second nearest neighbour, excluding self-matches) [44]. A mean synthetic-to-real DCR close to or exceeding this baseline indicates that synthetic samples are no closer to real records than real samples are to one another, providing evidence against memorisation. As a secondary check, any synthetic sample with DCR below a threshold of 0.05 was flagged as a potential near-copy. Unlike the classifier-based assessment above, DCR serves a descriptive role and was not included as an optimisation target in the composite score.

#### 3.4.6. Utility assessment

The final evaluation considered utility, reflecting this study’s goal of improving the prediction of rare events such as flight diversions through synthetic augmentation. Utility was assessed by comparing two scenarios: (1) Train-Real-Test-Real (TRTR), providing a baseline with only historical (real) data, and (2) Train-Augmented-Test-Real (TATR), incorporating synthetic diversions into training while testing solely on real data. When splitting the real dataset into training and testing sets, a stratified approach was used to preserve class balance. A Random Forest classifier was trained to predict flight diversions using the features listed in Table 1. The Random Forest was selected as a fixed evaluation classifier, consistent across all seven experiments, to ensure that any observed differences in utility scores are attributable to the augmentation framework rather than to classifier-specific behaviour. To prevent information leakage, time-related variables that could reveal actual arrival times were excluded, ensuring that the model relied solely on information available at the time of take-off. The use of only data available at departure time for the prediction task ensures operational applicability [45].

Because diverted flights represented only 127 out of 61,000 records, the dataset was highly imbalanced. To provide a robust evaluation of predictive performance under this class imbalance, we used the area under the precision-recall curve (PR-AUC) [46] and the normalised Matthews Correlation Coefficient (MCC) [47]. PR-AUC is particularly appropriate for rare-event prediction, as it focuses on the classifier's ability to identify the minority class while mitigating the misleading effects of high true-negative counts. MCC, a correlation coefficient between predicted and observed classifications, provides a balanced measure even in the presence of extreme class imbalance, capturing overall predictive quality beyond simple accuracy. This analysis was extended further by sampling varying numbers of synthetic diversions, ranging from 300 to 200,000 records, to assess the influence of augmentation size on the predictive utility. Comparing the TRTR and TATR scenarios with different augmentation sizes reveals important insights into the value of augmenting real data with synthetic diverted flights, particularly for improving the prediction of rare events such as flight diversions.

### 3.5. Multi-objective hyperparameter tuning

In this study, the principal methodological contribution lies in the design of a custom multi-objective function specifically crafted to enhance the quality of synthetic flight records, rather than merely the application of automated hyperparameter optimisation. This objective function was coupled with optimisation using the Tree-structured Parzen Estimator (TPE) algorithm from Akiba et al. [48]. Unlike classical Bayesian optimisation, which typically relies on Gaussian Processes, TPE constructs two non-parametric density estimators over the hyperparameter space: one for configurations associated with good performance and another for those with poor performance. New trials are sampled from regions where the ratio between these densities is maximised, guiding the search toward promising areas while maintaining exploratory diversity [49]. Compared to exhaustive grid search or uninformed random search, TPE is substantially more sample-efficient and scales effectively to high-dimensional and heterogeneous hyperparameter spaces, making it particularly suitable for deep generative models such as TVAE, CTGAN, and CopulaGAN, where the search space includes both continuous and categorical hyperparameters. In contrast, Gaussian Process–based Bayesian optimisation struggles with high dimensionality and categorical variables due to its reliance on kernel functions and cubic time complexity in the number of evaluations [50]. For computational reasons, the optimisation search was limited to 100 trials per model, focusing exclusively on the most influential hyperparameters to balance efficiency with practical resource constraints. It should be noted that, while the term multi-objective is used to reflect the simultaneous consideration of four quality dimensions, the optimisation is implemented via scalarisation: the four component scores are combined into a single composite score $\mathcal{S}$ (Eq. (4)), which TPE then maximises. As with any heuristic search, global optimality cannot be guaranteed; however, TPE's sample efficiency and the robustness of the selected trials across alternative weight configurations (Table 5) provide empirical support for the quality of the identified solutions.

The objective function was carefully designed to integrate multiple evaluation metrics into a single composite score, thereby aligning the optimisation process with the broader goal of generating high-quality synthetic data. Equal weighting was chosen as a principled default in the absence of domain evidence favouring any single criterion, and to avoid introducing an additional layer of tuning that would increase the risk of overfitting the optimisation procedure to a specific metric. Formally, the composite score $\mathcal{S}$ is defined as:

$$\mathcal{S} = w_1\,\mathcal{S}_{\text{real}} + w_2\,\mathcal{S}_{\text{sim}} + w_3\,\mathcal{S}_{\text{fid}} + w_4\,\mathcal{S}_{\text{util}} \tag{4}$$

where $\mathcal{S}_{\text{real}} \in [0,1]$ is the realism score, defined as the proportion of synthetic records with valid routes; $\mathcal{S}_{\text{sim}} \in [0,1]$ is the statistical similarity score, defined as the average of marginal and bivariate similarity scores; $\mathcal{S}_{\text{fid}} = 1 - \text{F1}_{\text{disc}} \in [0,1]$ is the fidelity score, defined as the complement of the cross-validated F1 score of a Random Forest classifier trained to discriminate real from synthetic records (rewarding indistinguishability); and $\mathcal{S}_{\text{util}} = \text{PR-AUC} \in [0,1]$ is the utility score, defined as the precision-recall AUC of a diversion classifier trained on augmented data and tested on real held-out records. All four components are bounded in $[0,1]$ and maximised jointly, with equal weights $w_1 = w_2 = w_3 = w_4 = 0.25$ used in this study.

F1 was chosen over balanced accuracy for the fidelity check because it considers both precision and recall, providing a more sensitive measure of the classifier's discriminative power. Balanced accuracy, while effective for handling class imbalance, can be less informative in this context: when the classifier is uncertain between real and synthetic data, it may still report a moderately high balanced accuracy (around 0.5), thereby underestimating the true fidelity of the synthetic data.

The PR-AUC was selected as the optimisation target over the MCC due to the strong class imbalance in the utility task—only 127 diversions out of 61,000 flights. PR-AUC specifically evaluates performance on the positive class (here, diverted flights), capturing the trade-off between precision and recall, which is critical when the positive class is rare. In contrast, MCC provides a single correlation-based metric that treats all classes equally; in highly imbalanced settings, it can be dominated by the majority class and may be less sensitive to improvements in detecting rare events. By maximising PR-AUC, the generative model was directly incentivised to produce synthetic diversions that enhance the classifier's ability to identify actual diversions, thereby improving utility in rare-event prediction.

Preliminary tests showed that increasing the depth of the encoder, decoder, generator, or discriminator beyond two layers degraded performance; hence, all architectures were fixed at two layers. At the same time, increasing the capacity of the initial layers (i.e., the number of neurons) improved performance. The optimisation focused on the most influential hyperparameters, with search ranges defined for each model. Table 4 summarises the hyperparameters and ranges explored. This systematic tuning improved the ability of the models to capture the underlying data structure while reducing issues such as mode collapse and overfitting.

**Table 4**
Summary of hyperparameters and search ranges used in the optimisation.

| Model | Hyperparameters and Search Ranges |
|---|---|
| TVAE | Number of epochs: [300, 6000]<br>Embedding dimension: [8, 600]<br>Neurons per layer (encoder/decoder): [8, 600]<br>Encoder/decoder depth: fixed at 2 |
| CTGAN<br>CopulaGAN | Number of epochs: [300, 6000]<br>Generator/discriminator learning rate: [1e-5, 1e-3]<br>Neurons per layer (generator/discriminator): [128, 512]<br>Generator/discriminator depth: fixed at 2 |

To verify the robustness of the equal-weight choice, a post-hoc sensitivity analysis was conducted by re-scoring all completed trials under five alternative weight configurations, without repeating the optimisation process: equal weighting (baseline), and four single-dimension-heavy schemes assigning a weight of 0.70 to each criterion in turn, with the remaining weight distributed equally among the other three. The results, summarised in Table 5, show that the selected trials remained largely stable across configurations. For TVAE, the selected trial changed only under the fidelity-heavy and realism-heavy schemes. For CTGAN, it changed under the fidelity-heavy and statistical-similarity-heavy schemes. For CopulaGAN, it changed under the fidelity-heavy and realism-heavy schemes. Notably, the utility-heavy configuration selected the same trial as the equal-weight baseline across all three models, suggesting that the baseline solution already provides near-optimal downstream utility. In all cases where the selected trial differed, the gain in the prioritised dimension came at the cost of a reduction in at least one other criterion—most notably utility, which dropped under the fidelity-heavy scheme across all three models—substantially so for TVAE and CTGAN (PR-AUC dropped from 0.44 to 0.26 and from 0.28 to 0.16, respectively), and more modestly for CopulaGAN (from 0.34 to 0.30). This suggests that the four criteria behave as partially competing objectives, and that equal weighting represents a principled and robust default that avoids privileging any single quality dimension.

**Table 5**
Sensitivity analysis of optimisation trial selection and individual quality scores under alternative weight configurations. $\mathcal{S}_{\text{real}}$: realism score; $\mathcal{S}_{\text{sim}}$: statistical similarity score; $\mathcal{S}_{\text{fid}}$: fidelity score; $\mathcal{S}_{\text{util}}$: utility score.

| Model | Weight scheme | Trial | $\mathcal{S}_{\text{real}}$ | $\mathcal{S}_{\text{sim}}$ | $\mathcal{S}_{\text{fid}}$ | $\mathcal{S}_{\text{util}}$ |
|---|---|---|---|---|---|---|
| | Equal (baseline) | 57 | 0.702 | 0.844 | 0.970 | 0.444 |
| | Utility-heavy | 57 | 0.702 | 0.844 | 0.970 | 0.444 |
| TVAE | Fidelity-heavy | 32 | 0.730 | 0.850 | 1.000 | 0.260 |
| | Statistical-heavy | 57 | 0.702 | 0.844 | 0.970 | 0.444 |
| | Realism-heavy | 3 | 0.783 | 0.836 | 0.898 | 0.329 |
| | Equal (baseline) | 33 | 0.841 | 0.832 | 0.888 | 0.277 |
| | Utility-heavy | 33 | 0.841 | 0.832 | 0.888 | 0.277 |
| CTGAN | Fidelity-heavy | 88 | 0.783 | 0.832 | 0.956 | 0.158 |
| | Statistical-heavy | 64 | 0.822 | 0.843 | 0.913 | 0.258 |
| | Realism-heavy | 33 | 0.841 | 0.832 | 0.888 | 0.277 |
| | Equal (baseline) | 42 | 0.862 | 0.850 | 0.942 | 0.335 |
| | Utility-heavy | 42 | 0.862 | 0.850 | 0.942 | 0.335 |
| CopulaGAN | Fidelity-heavy | 46 | 0.825 | 0.843 | 0.985 | 0.303 |
| | Statistical-heavy | 42 | 0.862 | 0.850 | 0.942 | 0.335 |
| | Realism-heavy | 81 | 0.874 | 0.849 | 0.928 | 0.332 |

## 4. Results

This section evaluates the quality of synthetic flight diversion records generated by the adapted models across the seven experiments described in Section 3.3. The aim is not to identify the single best model, but to demonstrate how our multi-objective hyperparameter tuning framework improves data quality compared with default configurations, and to assess the contribution of synthetic augmentation to the prediction of rare events such as flight diversions. The evaluation follows the framework presented in Section 3.4.

### 4.1. Realism assessment

Using the default configurations reported in Patki et al. [5] for the TVAE model resulted in a substantially lower realism score compared with the statistical GC. This outcome was expected, as deep learning models such as TVAE are inherently less effective when trained on very limited data (only 127 records of diverted flights in this case). By contrast, the CTGAN and CopulaGAN models, trained with default hyperparameters, achieved realism scores comparable to the GC model thanks to their conditional sampling mechanisms. Nevertheless, our multi-objective hyperparameter tuning substantially reduced the number of invalid flight routes. It led to a sevenfold improvement in the performance of the TVAE and doubled the realism scores of CTGAN and CopulaGAN models, as illustrated in Fig. 2.

To contextualise these scores in practical terms, the realism score represents the proportion of synthetic records containing origin–destination pairs that exist in the historical data, and therefore reflects whether the generated records correspond to realistic route connections. A score of 10.1% for the default TVAE means that approximately 9 out of 10 generated records contained route combinations with no historical precedent, rendering them unusable for augmentation. Similarly, the default CTGAN and CopulaGAN models produced valid routes in only around 4 out of 10 generated records. Since invalid routes are removed from the synthetic dataset before any subsequent evaluation step, a low realism score directly reduces the number of records available for augmentation. The sevenfold improvement to 70.2% for the optimised TVAE, and the approximate doubling to 84.0% and 86.2% for the optimised CTGAN and CopulaGAN respectively, therefore translate directly into substantially larger pools of usable synthetic diversion records. This directly expands the data available for training diversion prediction models.

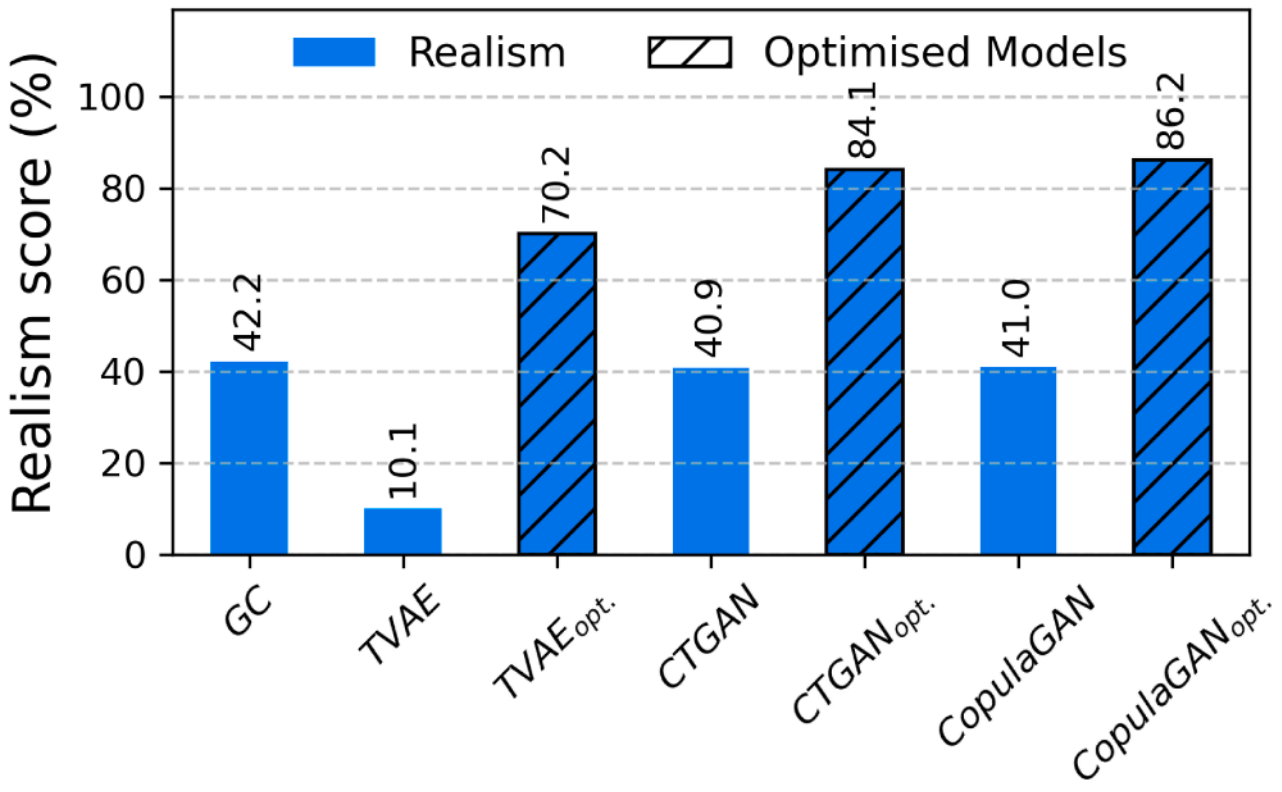

**Fig. 2.** Realism evaluation (Higher is better).

### 4.2. Diversity assessment

The multi-objective optimisation substantially improved the diversity of the synthetic data. All optimised models achieved broader coverage of the underlying patterns and clusters in the real dataset. As shown in Fig. 3, synthetic diversion records generated by the TVAE model pre-optimisation failed to capture the full variability of the real data, whereas those generated post-optimisation provided markedly better coverage.

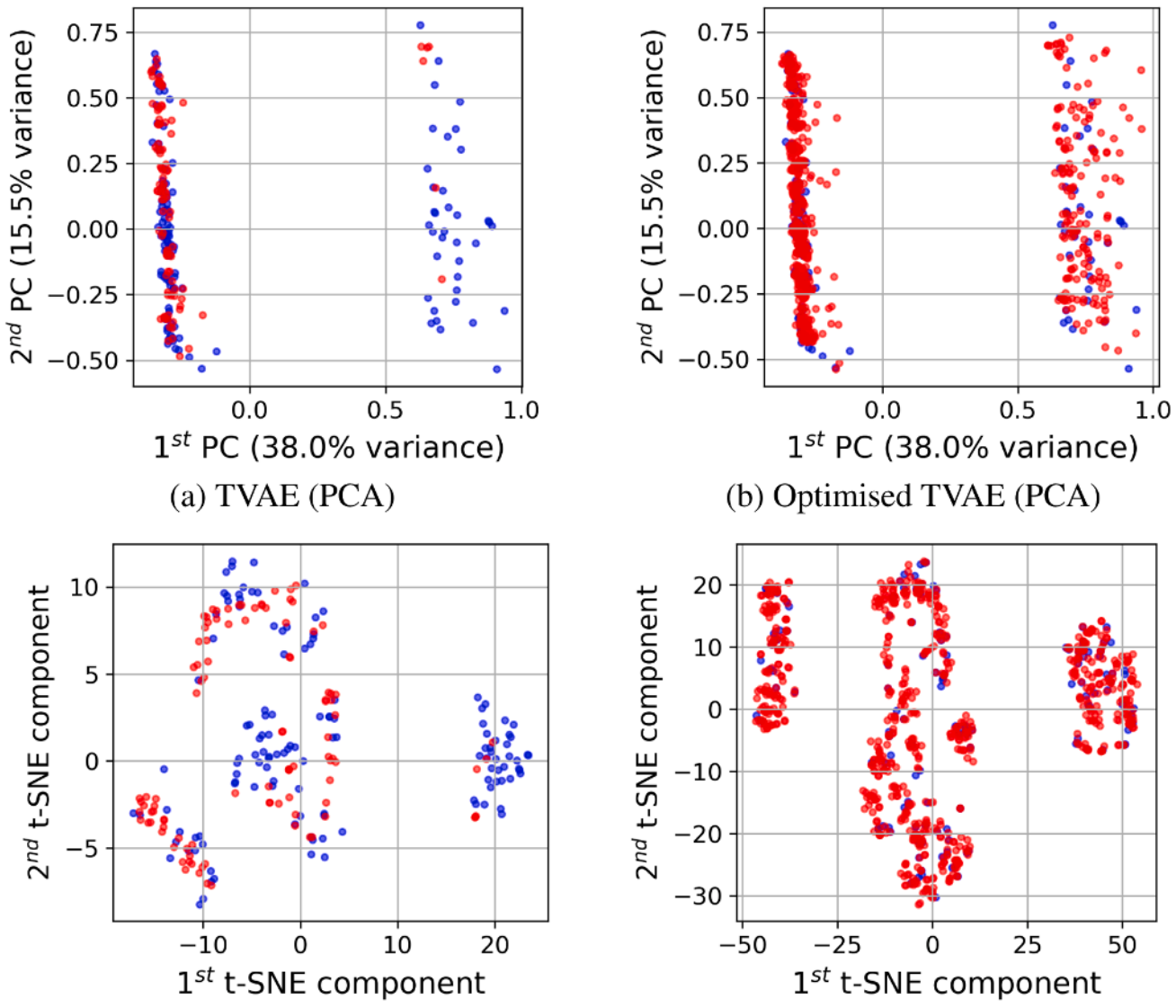

(a) TVAE (PCA)
(b) Optimised TVAE (PCA)
(c) TVAE (t-SNE)
(d) Optimised TVAE (t-SNE)

**Fig. 3.** Diversity comparison of real (blue) vs. synthetic (red) diversions generated by TVAE before (left) and after (right) the multi-objective optimisation. (For interpretation of the references to colour in this figure legend, the reader is referred to the web version of this article.)

Fig. 4 presents a comparison of class balance, illustrating the ratio of on-time to delayed flight departures in both the real and synthetic flight records. The pre-optimised CTGAN already exhibited a class balance close to that of the real data, owing to its conditional sampling mechanism, which allows it to accurately represent categorical values, leaving little room for improvement through optimisation. In contrast, the TVAE and CopulaGAN models benefited most from the optimisation.

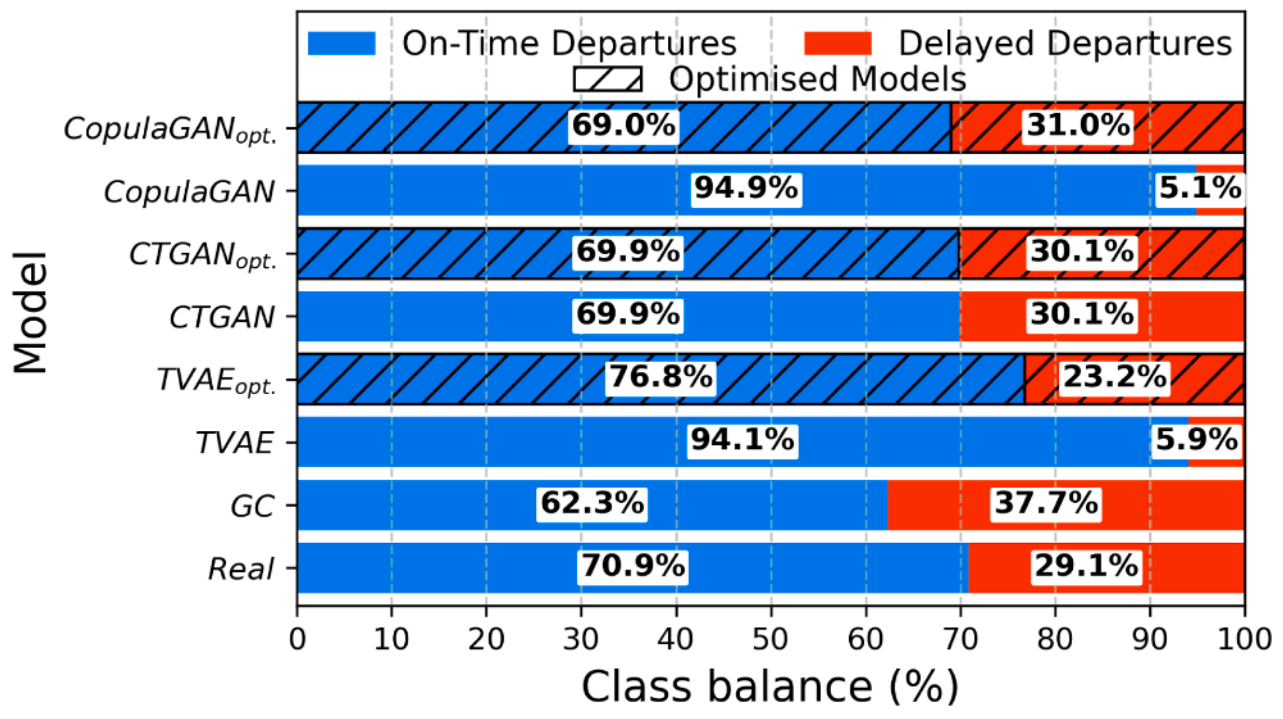

**Fig. 4.** Class balance between on-time (blue) and delayed (red) departures in real and synthetic diversion records generated by the different models. (For interpretation of the references to colour in this figure legend, the reader is referred to the web version of this article.)

### 4.3. Operational assessment

When examining the correlation between operational variables such as "Distance" and "Air Time", we observed that optimised models occasionally generated records deviating from real patterns. Fig. 5 illustrates this: synthetic records (red dots) are plotted on top of the real data (blue dots); synthetic records overlapping or close to real points are operationally feasible with respect to this correlation, while isolated red points indicate synthetic records that deviate from real patterns. The majority of synthetic records fall within the real data cloud, confirming that the optimised models capture the expected positive correlation between distance and air time. However, some isolated red points deviate noticeably from the real data cloud—for instance, synthetic records in the upper portion of both panels where air times far exceed the range observed in real data for the same distances. These anomalies indicate potential violations of operational feasibility, highlighting the need to refine the evaluation framework. This limitation arises because operational validity was included as a descriptive evaluation dimension rather than as a quantitative optimisation target; formalising it as a measurable score—for instance, by defining plausibility bounds based on flight envelopes or performance models—would allow it to be incorporated directly into the composite objective in future work. Incorporating such constraints into the evaluation framework can be accomplished with limited effort and would provide an effective way to more rigorously assess and enforce the operational validity of synthetic records.

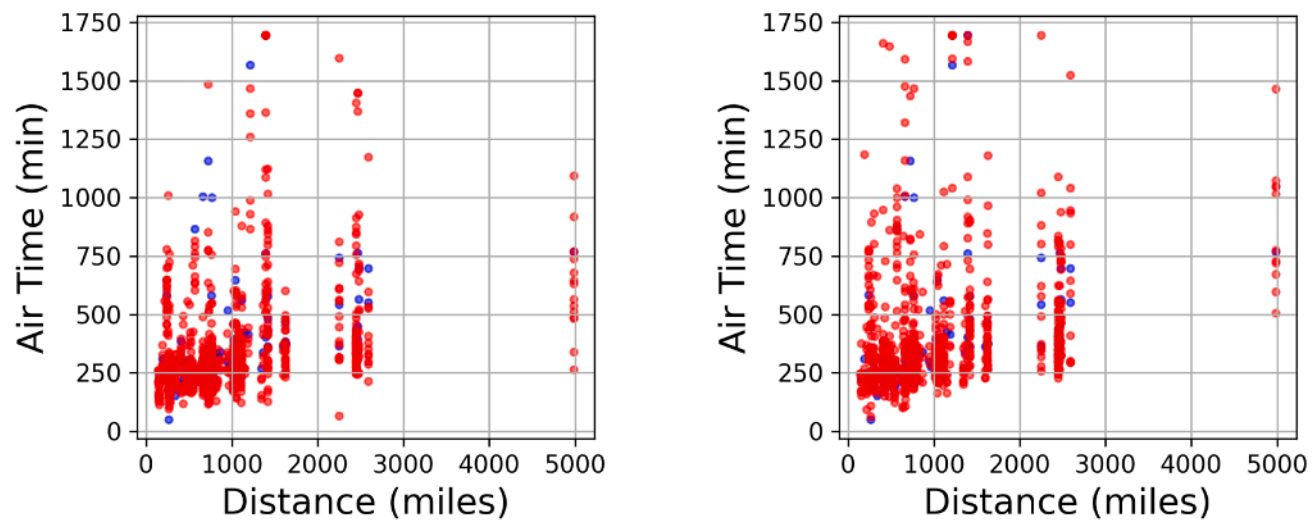

(a) Optimised CTGAN (b) Optimised CopulaGAN

**Fig. 5.** Operational correlation between air time and distance in real (blue) vs. synthetic (red) diversion records generated by the optimised CTGAN (a) and optimised CopulaGAN (b). (For interpretation of the references to colour in this figure legend, the reader is referred to the web version of this article.)

### 4.4. Statistical assessment

The optimisation process yielded synthetic data that more closely matched the statistical distributions of the historical records. Fig. 6 illustrates this improvement, showing marginal distributions of two features generated by CTGAN before and after optimisation, with the optimised model achieving greater statistical similarity to the real data.

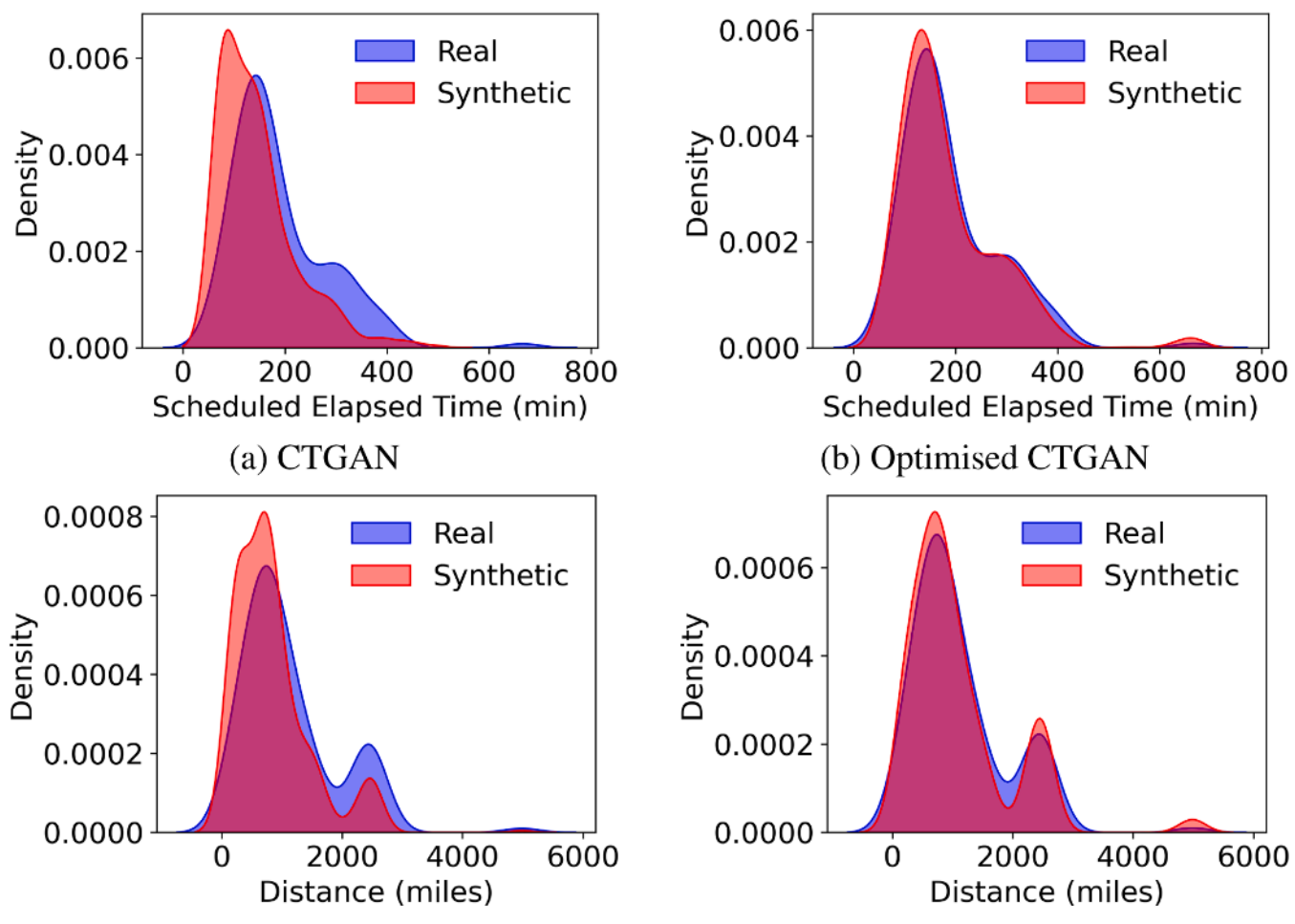

(a) CTGAN (b) Optimised CTGAN

(c) CTGAN (d) Optimised CTGAN

**Fig. 6.** Marginal distribution comparison of real (blue) vs. synthetic (red) features generated by CTGAN before (left) and after (right) the multi-objective optimisation. (For interpretation of the references to colour in this figure legend, the reader is referred to the web version of this article.)

The multi-objective function implemented in the tuning process aimed to maximise overall statistical similarity, which led to improvements in both marginal and bivariate similarities, as shown in Fig. 7. However, the gain in statistical similarity was slightly, though not substantially, greater than that achieved by the GC model, which employed KDE to estimate feature distributions and is therefore considered a benchmark for statistical similarity, particularly when compared with deep learning models trained on very small datasets.

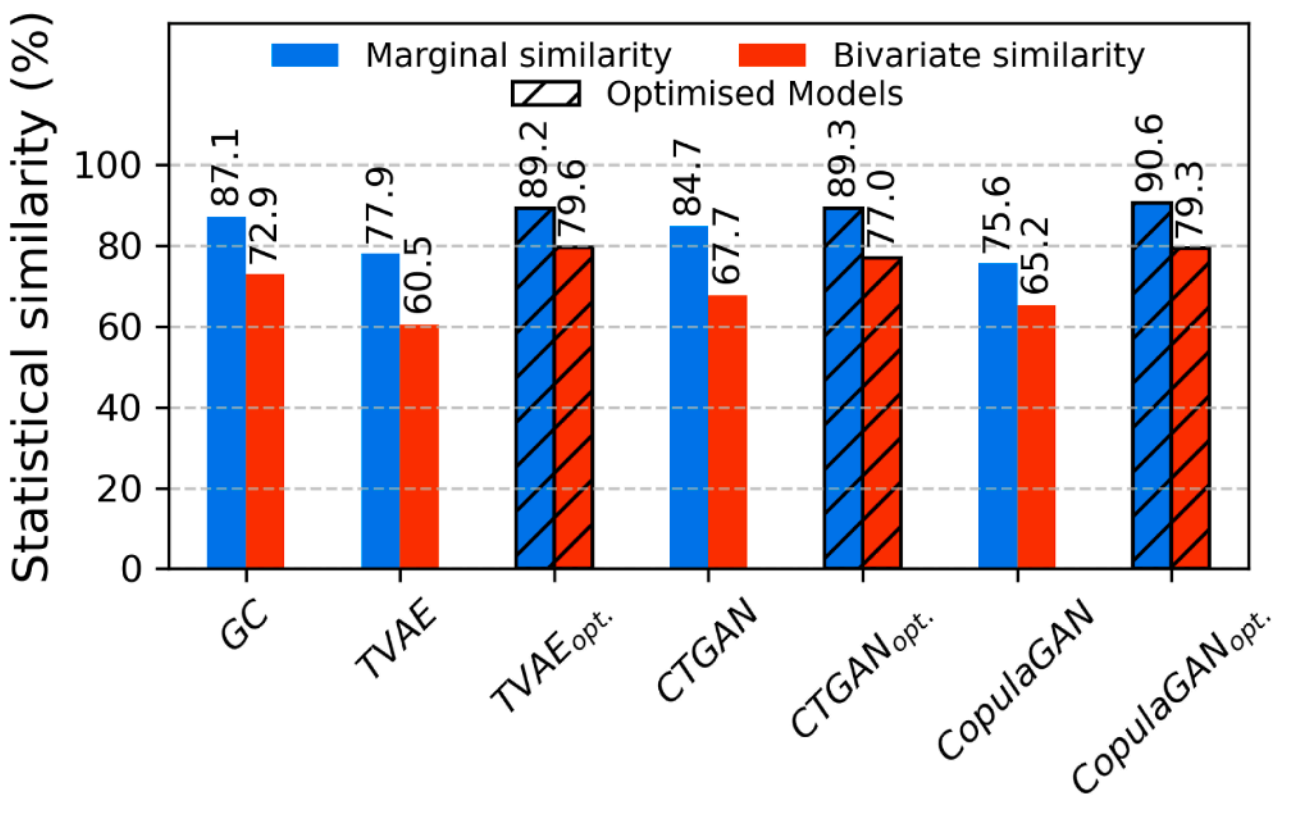

**Fig. 7.** Statistical evaluation (Higher is better).

### 4.5. Fidelity assessment

While the optimisation process in this study aimed to minimise the F1 score of the classifier distinguishing synthetic from real data, Fig. 8 shows reductions not only in the F1 score but also in the balanced accuracy, which remained around 0.5, as expected and discussed in Section 3.5. This demonstrates that the optimised models produced high-fidelity synthetic data that effectively confused the classifier, preventing it from reliably differentiating between real and synthetic records.

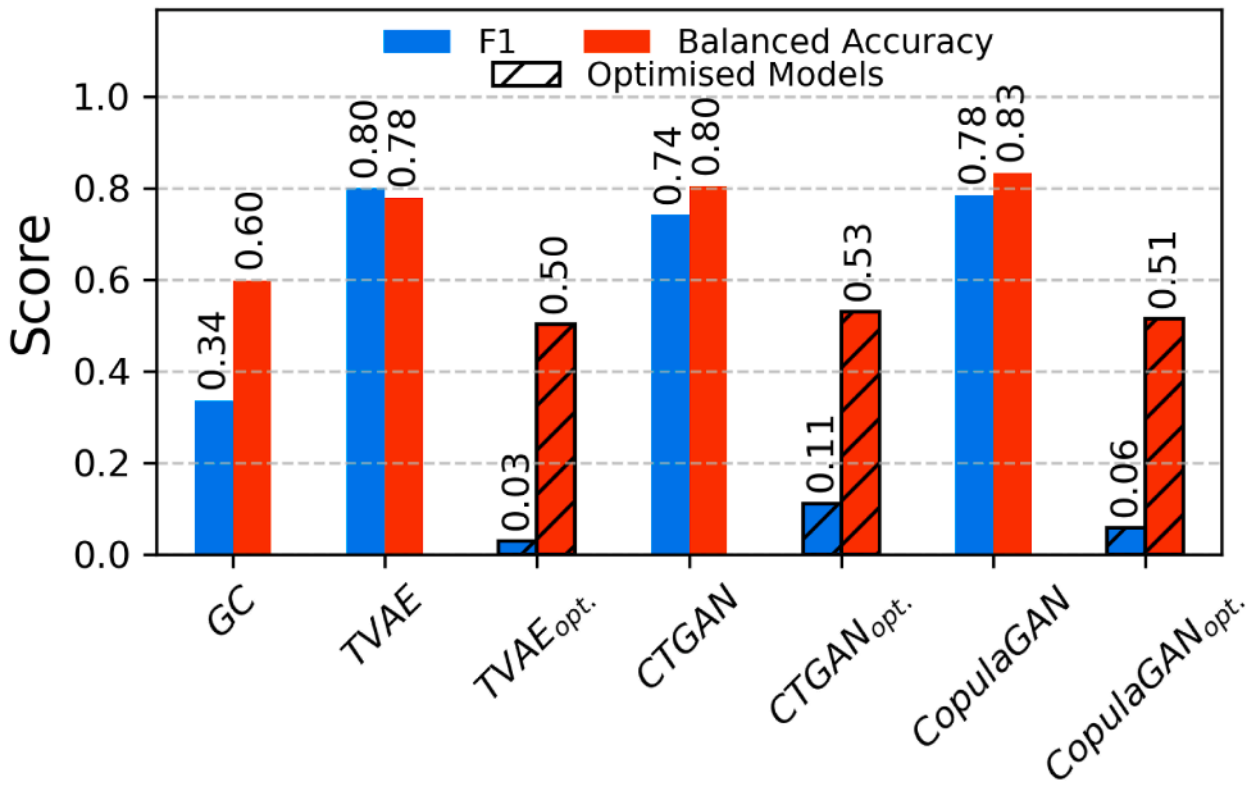

**Fig. 8.** Fidelity evaluation (Lower is better).

In addition to the indistinguishability results above, Table 6 reports the memorisation test results across all seven experiments. The real-to-real baseline mean DCR was 1.53, reflecting the natural spread of real diversion records in the normalised feature space. All models produced mean synthetic-to-real DCR values in the range 1.29–3.41, and no synthetic sample was flagged as a potential near-copy across any experiment. Five of the seven models yielded a DCR ratio greater than 1, indicating that synthetic samples were, on average, no closer to real records than real samples are to each other. The two TVAE configurations produced ratios slightly below 1 (0.84 and 0.87, respectively), suggesting a mild proximity to training samples; however, the minimum DCR values for these models (0.47 and 0.08) remained well above the memorisation threshold, and no samples were flagged. Taken together, these results provide no evidence of memorisation across any of the evaluated models.

**Table 6**
DCR results. The ratio is the mean synthetic-to-real DCR divided by the real-to-real baseline. A sample is flagged as a potential near-copy if its DCR < 0.05.

| Model | Mean DCR | DCR ratio | Flagged (%) |
|---|---|---|---|
| $GC$ | 1.86 | 1.22 | 0.00 |
| $TVAE$ | 1.29 | 0.84 | 0.00 |
| $TVAE_{opt.}$ | 1.33 | 0.87 | 0.00 |
| $CTGAN$ | 2.74 | 1.79 | 0.00 |
| $CTGAN_{opt.}$ | 1.68 | 1.10 | 0.00 |
| $CopulaGAN$ | 3.41 | 2.23 | 0.00 |
| $CopulaGAN_{opt.}$ | 1.54 | 1.01 | 0.00 |
| Real (baseline) | 1.53 | 1.00 | — |

### 4.6. Utility assessment

The multi-objective tuning process was designed to maximise the PR-AUC of the classifier predicting flight diversions. As shown in Fig. 9, this optimisation led to notable improvements not only in PR-AUC but also in MCC, indicating that the generated synthetic data enhanced the classifier's ability to predict rare diversion events.

The utility scores of models trained solely on real data in the Train-Real-Test-Real (TRTR) scenario—represented by the horizontal lines—were expectedly low, reflecting the scarcity of flight diversions in the historical records. Despite the absence of features strongly correlated with diversions—such as weather conditions, air traffic congestion, crew availability, aircraft maintenance status, and operational delays—the utility scores of optimised models trained on augmented data and tested on real data (TATR) were substantially higher than those of pre-optimised models and, most importantly, exceeded the scores of models trained exclusively on real data in the TRTR scenario.

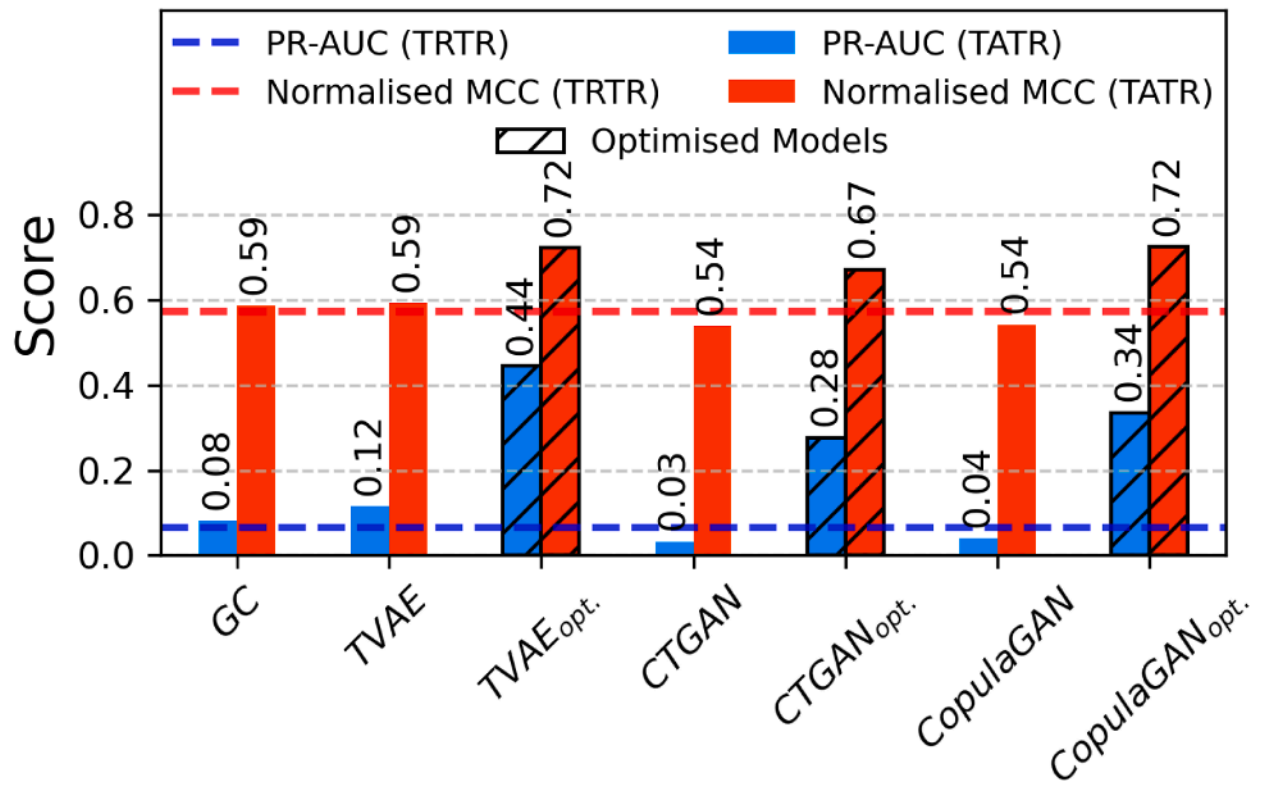

**Fig. 9.** Utility evaluation (Higher is better).

This demonstrates the effectiveness of our optimisation framework in generating high-utility synthetic data and provides clear evidence that augmenting real data with synthetic instances of the minority class can significantly enhance predictive performance for rare events such as flight diversions.

While the results reported above were obtained using an augmentation size of 1,000, Fig. 10 shows how predictive utility varies with augmentation size. Synthetic augmentation results are shown as circular markers, with the smoothed LOWESS trends represented by the lines [51].

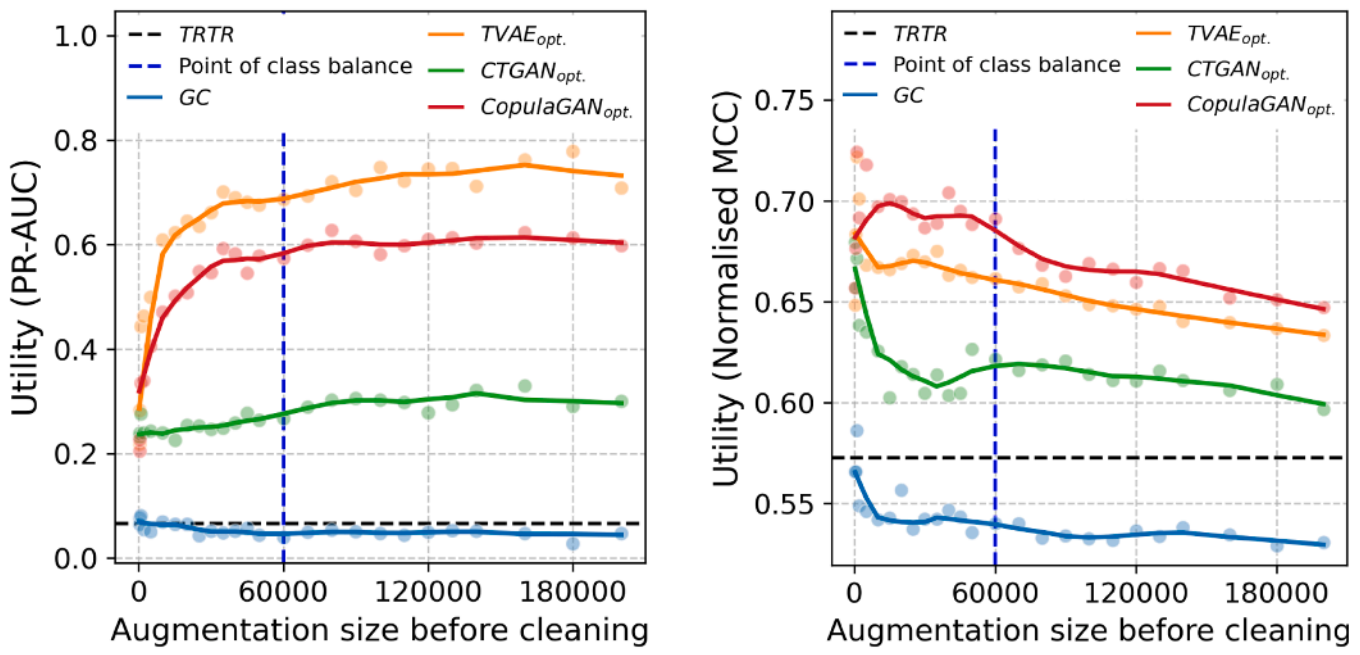

(a) PR-AUC (b) MCC

**Fig. 10.** Impact of augmentation size on utility scores.

For the statistical GC model, both utility scores generally remained below those of the Train-Real-Test-Real (TRTR) baseline. In contrast, synthetic data generated by the optimised TVAE and CopulaGAN models showed PR-AUC values that rose rapidly with increasing augmentation size before tapering off near the point of class balance (indicated by the vertical blue line), where the number of diverted and non-diverted flights becomes equal. The optimised CTGAN displayed a similar pattern, yet with a shallower slope. The tendency of PR-AUC to level off warrants further investigation. In particular, improvements to the post-generation cleaning procedure are needed to ensure that larger augmentations do not introduce noisy data. This will help determine whether the class balance point represents a genuine characteristic threshold or merely an artefact of data quality. For MCC, the optimised models achieved higher values than the TRTR baseline, but the scores declined as the augmentation size increased.

The observed behaviour of PR-AUC and MCC can be attributed to their differing sensitivities to class imbalance. As the augmentation size increases, the classifier benefits from additional examples of the minor-ity class, improving its ability to detect diversions and thereby raising PR-AUC. However, heavy oversampling also shifts the class distribution and increases the likelihood of false positives. While PR-AUC rewards the improved detection of rare events, MCC penalises errors across both classes equally; thus, the rise in false positives causes MCC to decline even as PR-AUC improves. This highlights the importance of selecting an optimal augmentation size, where the benefits of better rare-event detection are maximised without introducing excessive false positives that degrade overall model performance.

## 5. Discussion

The experimental results demonstrate that multi-objective hyperparameter optimisation consistently improved synthetic data quality across all six evaluation dimensions relative to default configurations. Realism improved most dramatically for the TVAE, increasing sevenfold (from 10.1% to 70.2%), while the CTGAN and CopulaGAN roughly doubled their valid-route proportions (to 84.0% and 86.2%, respectively). Statistical similarity gains were substantial for the TVAE and CopulaGAN, and moderate for the CTGAN, which already achieved competitive similarity under default settings owing to its conditional sampling mechanism operating directly on the original categorical distributions—unlike CopulaGAN, which trains on a Gaussian copula transformation of the data. Fidelity improved markedly for all three optimised models, with the discriminative classifier approaching random performance (F1 near zero), indicating that optimised synthetic records were effectively indistinguishable from real ones, while the DCR memorisation test confirmed that no synthetic sample was flagged as a near-copy across any experiment. Most importantly, utility improved in a practically significant way: optimised models yielded TATR PR-AUC scores of 0.44, 0.28, and 0.34 for the TVAE, CTGAN, and CopulaGAN, respectively, compared with a TRTR baseline of 0.066—representing gains of more than sixfold over training on real data alone.

These results underscore the importance of adopting a multifaceted evaluation approach rather than relying on a single metric. The CTGAN already demonstrated class balance close to that of the real data prior to tuning—thanks to its conditional sampling mechanism—yet the multi-objective optimisation process enhanced its performance in other critical areas, including realism, statistical similarity, fidelity, and utility. Conversely, the statistical GC model, despite achieving strong statistical similarity through kernel density estimation, consistently underperformed on utility, confirming that distributional similarity alone is insufficient for downstream prediction and that the explicit inclusion of a utility criterion in the composite objective is essential.

The model analysis underscored the critical influence of model architecture on generating high-quality synthetic data. Specifically, increasing the depth of the generative models (i.e., adding more layers) generally reduced performance, whereas increasing the capacity of the initial layers (i.e., the number of neurons) improved it. This finding highlights the importance of carefully balancing depth and capacity, particularly when working with relatively small datasets, as architectural choices directly affect the models' ability to capture complex data distributions.

Beyond architectural choices, optimisation improved the plausibility of generated records and substantially reduced the number of invalid routes. However, the realism assessment was limited to identifying airport connections absent from historical data. Expanding the assessment to include checks on airport-specific features—such as ensuring synthetic Taxi In and Taxi Out times closely match real distributions for the same airports—would help ensure that the generated records are realistic and operationally meaningful.

The operational feasibility assessment offered an additional perspective on model behaviour. While visual comparisons of operational correlations provided useful information, they were insufficient for rigorously assessing whether generated points were operationally valid. Notably, some synthetic records exhibited air times that exceeded the range observed in real data for the same distances, indicating potential violations of physical plausibility. Formalising operational validity as a quantitative score—for instance, by defining plausibility bounds derived from flight envelopes or aircraft performance models such as BADA—would allow it to be incorporated directly into the composite objective, complementing the current descriptive assessment.

The fidelity assessment highlighted the importance of selecting appropriate optimisation targets. Minimising the F1 score rather than balanced accuracy proved effective, as balanced accuracy tends to stabilise around 0.5 when classifiers are uncertain, masking the true fidelity of the synthetic data. In contrast, the F1 score—by considering both precision and recall—provided a more sensitive measure of classifier confusion, ensuring that the optimisation process successfully reduced distinguishability between real and synthetic records.

Beyond fidelity, the utility results revealed that both the optimisation metric and the augmentation size should be aligned with the intended application when tuning generative models. If the goal is to maximise detection of rare diversion events—where high recall is critical, such as in safety-sensitive contexts—prioritising PR-AUC and using larger augmentation sizes is advantageous. Conversely, when balanced predictions are required across diverted and non-diverted flights—for example, to avoid overwhelming air traffic controllers or airline operators with false alarms—MCC should be prioritised, and more moderate augmentation sizes used. In practice, the choice between PR-AUC and MCC as the optimisation target reflects a trade-off between maximising rare-event detection and maintaining overall predictive balance. Researchers and practitioners should therefore tailor their optimisation objectives to the operational priorities of their specific use case.

From an operational standpoint, the proposed framework can be integrated into an air traffic management decision-support pipeline in a straightforward manner. The augmentation step is performed entirely offline: historical diversion records are used to train and optimise the generative model, synthetic diversion records are produced, and the augmented dataset is used to train a diversion risk classifier. At inference time, the classifier operates on tabular features routinely available at departure—including carrier, aircraft type, origin–destination pair, scheduled departure time, and departure delay—and produces a binary diversion prediction for each flight. This output could be used by airline dispatchers or network managers to prioritise pre-departure monitoring, trigger early coordination with alternate destination airports, optimise fuel uplift planning, or adjust ATFM slot allocations for flagged flights. Because the classifier requires no real-time trajectory data, it is compatible with pre-departure tools such as flight plan processing systems and network management platforms. As new diversion records accumulate over time, the generative model can be periodically retrained to reflect evolving operational patterns, making the framework suitable for deployment in a continuous improvement cycle.

The present study has several limitations that warrant acknowledgement. The dataset was restricted to one month of U.S. domestic flight records involving New York airports, yielding 127 diverted flights out of approximately 61,000 records. While this scope was sufficient to demonstrate the effectiveness of the proposed framework, the limited temporal and geographical coverage constrains the generalisability of the *trained* generative models. Deep generative models trained on a narrow operational window may not capture the full variability of diversion patterns across seasons, regions, or airline types. This limitation applies to the trained model instances rather than to the general optimisation approach. While the six-stage evaluation framework was designed for aviation tabular records and some of its dimensions—such as route realism and operational validity—are domain-specific, the underlying principle of jointly optimising multiple complementary quality criteria for minority-class augmentation is transferable to other domains with appropriate adaptation of the evaluation criteria. Replication across broader datasets would nonetheless be necessary before the trained generative models could be considered representative of the wider diversion population.

The dominant computational expense lies in the multi-objective optimisation loop rather than in generative model training or inference. In the present experiments, the optimisation was capped at 100 trials per model due to resource constraints; each trial required training a deep generative model on 127 records and evaluating the composite objective on a standard CPU-only machine (Intel Core i7-1185G7, 16 GB RAM). Generating 1,000 synthetic records from the trained model is substantially faster and negligible in comparison. In terms of scalability, the framework is well-suited to the small-data regime that characterises rare aviation events: as the number of real minority-class records grows, generative model training becomes more stable and data-efficient. The principal scalability consideration is the number of optimisation trials required, which grows with the dimensionality of the hyperparameter search space; however, the TPE algorithm handles this more gracefully than grid search, as its sample efficiency improves relative to exhaustive methods as dimensionality increases [49].

## 6. Conclusion & future work

This study adapted several generative models to produce synthetic records of flight diversions, which were then used to augment this minority class in historical datasets and improve the training of diversion prediction models. To this end, we developed a multi-objective optimisation framework for hyperparameter tuning tailored to this use case. The results demonstrated substantial improvements in the quality of synthetic data for optimised models compared with those trained using default configurations.

Across the case study, the TVAE benefited most from optimisation in terms of realism, while the GAN-based models—which already achieved competitive realism and class balance under default configurations owing to their conditional sampling mechanisms—gained markedly in fidelity and utility. Notably, the statistical GC model, despite achieving strong distributional similarity, consistently underperformed on utility, confirming that statistical faithfulness alone does not translate into predictive value and that the explicit inclusion of a utility criterion in the composite objective is essential. Across all optimised models, augmenting the training data with synthetic diversion records improved diversion prediction over training on real data alone.

The study provides clear evidence of the effectiveness of synthetic augmentation in enhancing rare-event prediction within the air transportation domain. While flight diversions were used as an illustrative example, the underlying principle of jointly optimising multiple complementary quality criteria for minority-class augmentation is transferable to other domains, provided that the evaluation dimensions are adapted to reflect domain-specific requirements.

Future research could further explore using the Matthews Correlation Coefficient (MCC), or the average of MCC and PR-AUC, as optimisation objectives for the utility assessment instead of relying solely on PR-AUC. The present experiments were limited to 100 optimisation trials due to computational constraints; extending both the search space and the number of trials would allow for a more exhaustive exploration of hyperparameters. A rigorous experimental comparison between the proposed multi-objective generative augmentation framework and traditional resampling approaches such as SMOTE, evaluated across the same six quality dimensions employed in this study, would further contextualise the contribution of generative augmentation relative to conventional imbalance-handling techniques and represents a promising direction for future work. Furthermore, the utility evaluation in this study relied on a single classifier; investigating whether the observed augmentation gains generalise across different downstream model families—and whether more effective classifier-augmentation combinations exist, for instance between tree-based models and specific generative architectures—would be a valuable complement to the present work. Formalising operational validity as a quantitative score—for instance, by defining plausibility bounds based on flight envelopes or aircraft performance models—would allow it to be incorporated directly into the composite objective $\mathcal{S}$, complementing the current descriptive assessment. Expanding the study to cover multiple months and a wider range of airports and airline operators would address the current limitations in temporal and geographical coverage, enabling assessment of whether the generative models trained on a single operational context remain effective under more diverse conditions. Looking further ahead, integrating the proposed framework into an operational air traffic management decision-support pipeline—including automated refresh cycles as new diversion records accumulate over time—would be a natural next step toward operational deployment.

## CRediT authorship contribution statement

**Karim Aly:** Writing – original draft, Visualization, Validation, Software, Methodology, Investigation, Formal analysis, Data curation, Conceptualization; **Alexei Sharpanskykh:** Writing – review & editing, Supervision, Project administration, Methodology, Funding acquisition, Conceptualization; **Jacco Hoekstra:** Writing – review & editing, Supervision.

## Funding

This paper is based on the work done in the SynthAIr project. SynthAIr has received funding from the SESAR Joint Undertaking under the European Union's Horizon Europe research and innovation programme under grant agreement No 101114847. Views and opinions expressed are however those of the authors only and do not necessarily reflect those of the European Union or SESAR 3 Joint Undertaking. Neither the European Union nor SESAR 3 Joint Undertaking can be held responsible for them.

## Declaration of competing interest

The authors declare that they have no known competing financial interests or personal relationships that could have appeared to influence the work reported in this paper.

## Acknowledgements

The authors would like to acknowledge the contributions and support of the SynthAIr consortium in the development of this work.

## Data availability

The code and data used in this study are publicly available at: https://github.com/SynthAIr/SynFlyDiv.